\documentclass[11pt,a4paper]{article}
\usepackage[hyperref]{emnlp2020}
\usepackage{times}
\usepackage{latexsym}

\usepackage{microtype}
\usepackage{graphicx}

\newcommand\blfootnote[1]{%
  \begingroup
  \renewcommand\thefootnote{}\footnote{#1}%
  \addtocounter{footnote}{-1}%
  \endgroup
}

\aclfinalcopy % Uncomment this line for the final submission
\title{DSC-IIT ISM at WNUT-2020 Task 2: Detection of COVID-19 informative tweets using RoBERTa
}

\author{
\\
  \\
  
  \And
  Sirigireddy Dhanalaxmi, Rohit Agarwal* and Aman Sinha*\\
  \\
  Indian Institute of Technology (Indian School of Mines) Dhanbad, India \\
  {\tt \{sirigireddydhanalaxmi, agarwal.102497, amansinha091\}@gmail.com} \\
  \And
  \\
  \\
  \\} 

\date{}

\begin{document}
\maketitle

% ////////////////////////////////////////////////////////////////////
\begin{abstract}
Social media such as Twitter is a hotspot of user-generated information. In this ongoing Covid-19 pandemic, there has been an abundance of data on social media which can be classified as informative and uninformative content. In this paper, we present our work to detect informative Covid-19 English tweets using RoBERTa model as a part of the W-NUT workshop 2020. We show the efficacy of our model on a public dataset with an F1-score of 0.89 on the  validation dataset and 0.87 on the leaderboard.
\end{abstract}

% ////////////////////////////////////////////////////////////////////
\section{Introduction}
\begin{figure}[t]
\centering
\includegraphics[width=7cm]{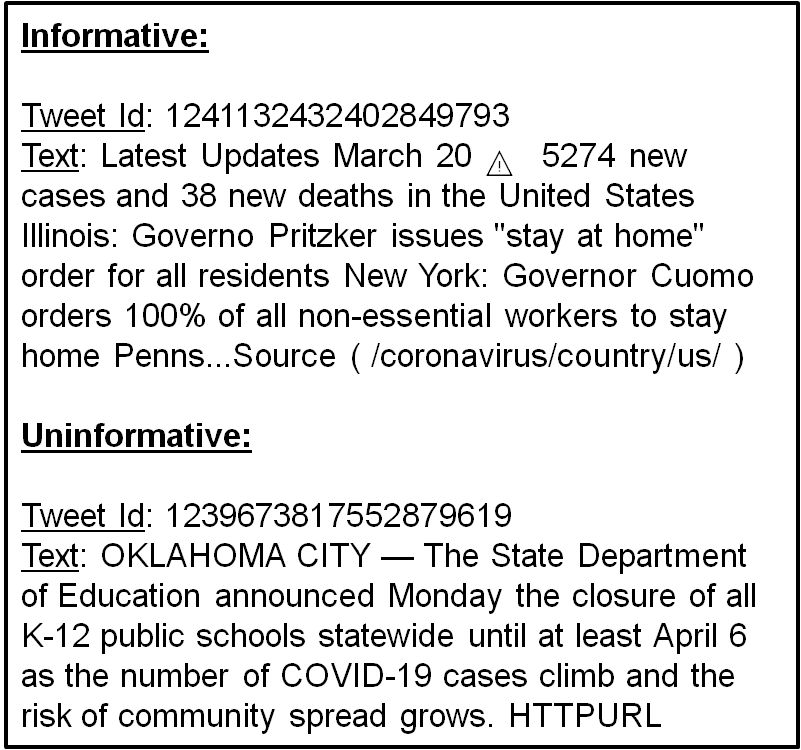}
\caption{An example of informative and uninformative tweet.}
\label{fig-1}
\end{figure}

Text analysis of social media data gives broader
insights into various topics discussed among
people. Twitter is a social media platform where people interact through short texts.
This paper constitutes our work for the Shared Task 2 of the 6th Workshop on Noisy User-generated Text (W-NUT) \citep{covid19tweet} where we need to classify the Covid-19 English tweets as informative or uninformative. In the context of this shared task, a tweet is considered informative if it is about recovered, suspected, confirmed, and death cases and location or travel history of the cases, and all the other tweets fall into the category of uninformative class. Figure \ref{fig-1} shows an example of both informative and uninformative tweets.

We applied various machine learning models such as logistic regression, Naive Bayes, random forest classifier, support vector machine (SVM), and multi-layer perceptron (MLP).
We have also used several state-of-the-art architectures like BERT, DistilBERT, RoBERTa, and ALBERT for detecting informative tweets.
We provide a comparative study of all these models and found the RoBERTa model to perform best among all the models.

\blfootnote{
    \textbf{*} Equal contribution.\\
}

This paper's outline is as follows: Section \ref{sec-2} discusses the previous works related to our paper. Section \ref{sec-3} describes the dataset and the data preprocessing steps. 
Section \ref{sec-4} describes our methods and section \ref{sec-5} discusses the implementation details of our approaches. Section \ref{sec-6} contains the analysis of the results, which is followed by the conclusion in section \ref{sec-7}.

% ////////////////////////////////////////////////////////////////////
\section{Related Work}
\label{sec-2}

Detection of useful-crisis-related content has been pivoting around Twitter due to its interactive media via microtexts \citep{martinez2018twitter}. Continuous Bag-of-Words (CBoW) based approach has been used for text classification \citep{sriram2010short}. \newcite{10.1145/1963405.1963500} proposes the use of different user-based features representing messages and tweet propagation for classifying tweet credibility.

Some works proposed use of SVM \citep{DBLP:journals/corr/abs-1803-05495}, logistic regression \citep{article}, random forest classifier \citep{article1} and word embedding based method \citep{DBLP:journals/corr/BadjatiyaG0V17} for classification of tweet contents.  \newcite{8258082} provided the use of unsupervised methods to cluster news topics from tweets. 

The capability of dependency learning and semantic information extraction enables us to learn complex decision boundaries. The evolution around capturing the semantic relationships between words lead to the widely used Transformer \citep{vaswani2017attention} architecture.

Such models have outperformed conventional methods over natural language processing (NLP) tasks. They have been widely used for various real-world applications such as language modeling \citep{DBLP:journals/corr/abs-1904-09408}, sarcasm detection \citep{kumar-jena-etal-2020-c}, summarization \citep{egonmwan-chali-2019-transformer}, and other language tasks. 

%//////////////////////////////////////////
\begin{figure}[t]
\centering
\includegraphics[width=7cm]{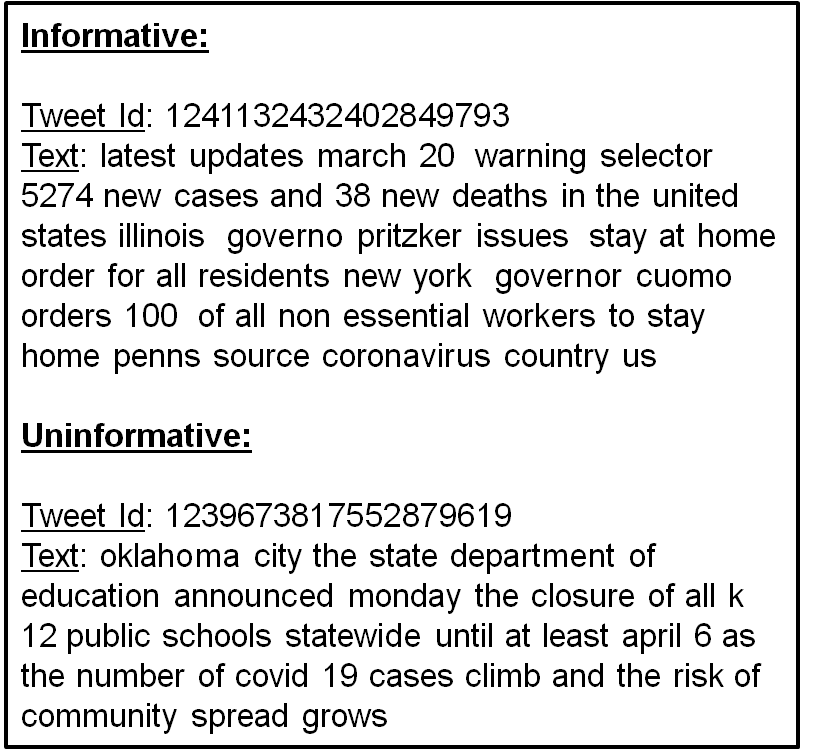}
\caption{Preprocessed data of the examples shown in fig.\ref{fig-1}.}
\label{fig-2}
\end{figure}

\section{Dataset}
\label{sec-3}

The training and validation data consists of 7000 and 1000 samples, respectively. 
The test dataset consists of 12000 tweets, out of which 2000 tweets were selected by the organizers for final evaluation.
The actual labels of the test dataset was not revealed by the shared task, hence the accuracy metric is only reported for the validation dataset in this paper.
The number of samples in each class is presented in Table \ref{tab-1}. The maximum, minimum, and average word count of the train and validation data is also shown in Table \ref{tab-1}.

\paragraph{Preprocessing data}  All the texts are converted to lower-case, and the emojis are replaced by their corresponding textual description. 
Further, contractions in the texts are fixed, and URLs and non-ascii characters are removed. 
It can be seen from Table \ref{tab-1} that the range of word count has increased after preprocessing for both the train and validation data. The word count increased due to emojis' conversion to text and decreased due to the removal of non-ascii characters. An example of a preprocessed tweet for both the informative and uninformative class is shown in Figure \ref{fig-2}.

\begin{table}
\centering
\begin{tabular}{lrl}
\hline \textbf{} & \textbf{Train} & \textbf{Validation} \\ \hline

\multicolumn{3}{c}{\textit{Number of samples in each class}} \\ 
\hline
Informative & 3303 & 472 \\
Uninformative & 3697 & 528 \\
\hline

\multicolumn{3}{c}{\textit{Word count - before preprocessing}} \\ 
\hline
 Maximum & 76 & 62 \\
 Minimum & 8 &  11 \\
 Average & 35.87 & 37.052 \\
\hline

\multicolumn{3}{c}{\textit{Word count - after preprocessing}} \\ 
\hline
 Maximum & 217 & 69\\
 Minimum & 7 & 10\\
 Average & 36.301 & 37.215\\
\hline
\end{tabular}
\caption{\label{tab-1} Dataset statistics - number of samples in each classes, and word count before and after preprocessing data}
\end{table}

% ////////////////////////////////////////////////////////////////////
\section{Methods}
\label{sec-4}

We have applied various conventional machine learning and transformer-based approaches.

\subsection{Conventional approaches}
We used traditional ways of word representation such as Bag-of-Words (BoW) and TF-IDF for detecting informative tweets using classifiers such as logistic regression, SVM, Naive Bayes, random forest classifier, and 2-layer MLP.

\subsection{Transformer based approaches}
Transformer is a way of improving the performance of NLP models. It is an encoder-decoder-type architecture that observes the whole of the input sequence at once.  Unlike the recurrent sequential method, it uses an attention mechanism to detect long term dependencies. In this paper, we focus on experimenting with transformer-based architectures like BERT, DistilBERT, RoBERTa, and ALBERT.

\paragraph{BERT} \newcite{devlin2018bert} presents a bi-directional transformer-based language model, pre-trained on deep bidirectional representations from unlabeled text. It is jointly conditioned on both left and right context in all layers, and it is known for outperforming several state-of-the-art systems for various NLP tasks. We have used BERT-base-uncased pre-trained model to perform the informative tweet classification.

\paragraph{DistilBERT} \newcite{sanh2019distilbert} proposes an approximate version of BERT using half the number of parameters. It improves the inference time while retaining 97\% of the performance of BERT. The pre-trained model we used is DistilBERT-base-uncased to analyze the comparative classification with respect to the BERT model.

\paragraph{RoBERTa}\newcite{liu2019roberta} adopts the training mechanism used by BERT with a significantly longer training time over longer sequences. It differs from BERT as it uses a dynamic masking pattern compared to static in prior. We have used RoBERTa-base pre-trained model. It is trained with a significantly large dataset and outperforms BERT, DistilBERT, and other variants for various downstream tasks.

\paragraph{ALBERT} \newcite{lan2019albert} introduces another light version of BERT, with low memory consumption and high training speed by which it outperforms the state-of-the-art models for various benchmark datasets. In our experiment, we used the pre-trained ALBERT-base-v1 model.

We have used all these pretrained models with the same parameters (discussed in section 5.2) except for RoBERTa and DistilBERT, which required small changes. In RoBERTa, before tokenizing the sentences, prefix space has to be set true, along with the addition of special tokens. In DistilBERT, the token type id's are not considered.

%%%%%%%%     Table-2 %%%%%%%%%%%%%%%%

\begin{table}
\centering

\begin{tabular}{lp{1.5cm}p{1.5cm}}
\hline \textbf{Classifier} & \textbf{Bag-of-Words} & \textbf{TF-IDF Vectors} \\ 
\hline
Logistic Regression  & 0.78318 & 0.78331 \\
SVM & 0.78054 & 0.78472 \\
Naive Bayes & 0.76371 & 0.74449 \\
Random Forest & 0.55489 & 0.56447 \\
MLP & 0.78695 & \textbf{0.79912} \\
\hline
\end{tabular}
\caption{\label{tab-2} F1 score of conventional approaches. }

\end{table}

% ////////////////////////////////////////////////////////////////////

\section{Implementation}
\label{sec-5}

\subsection{Conventional approach}
The BoW and TF-IDF vectors are obtained after the given training, and validation sets undergo preprocessing procedure. The liblinear solver is used in the logistic regression. The maximum depth of the decision tree in random forest classifier is set as 8. In the MLP classifier, lbfgs solver is used with alpha value set as 1e-5, the number of hidden layers is 2 with 5 and 2 neurons in the first and second layers, respectively. Other parameters concerning the conventional methods use default values.

\subsection{Transformer-based approach}
\label{sec-5.2}
The sentences after undergoing the cleaning process are tokenized using the pre-trained model’s tokenizer. Special tokens are added to detect the start and end of a sentence, and each token is mapped with an id. Next, the padding layer is added with value 0 and truncated to a maximum length of 100 to maintain equal lengths of the embeddings. Attention masks are used to detect padded tokens and actual words. Mask is set to 0 if the token id is 0, else it is set to 1.

The actual training set is further split into two parts (9:1 ratio), i.e., train and dev set to check which learning rate the model performs better. 
The input arguments are passed to evaluate our validation dataset. Finally, the F1 score is calculated between predicted and actual labels of the validation set.

\paragraph{Reproducibility} We have considered batch size as 32, the learning rate of the optimizer as 2e-5, and its epsilon value is set as 1e-8. We trained our model for 4 epochs as determined by optimising on the dev set. To get the reproducible results, we set the seed value for all the Python packages. The torch seed, manual seed, and NumPy seed are set as 0. Further, while using CuDNN backend, we set deterministic as true and benchmark as false.

%%%%%%%%%%%%%%%%%%%%%%%%%%%%%%%%%%%%%%%%%%%%%%

\section{Results}
\label{sec-6}

The F1 scores on the validation dataset obtained for conventional methods and transformer-based methods are presented in Table \ref{tab-2} and Table \ref{tab-3} respectively. 

It is observed that the TF-IDF vectorizer gives better results when compared to BoW in almost all the conventional approaches. Among all the conventional approaches, MLP gives the best result.

The transformer-based methods performed better compared to all the conventional approaches. BERT and RoBERTa showed competitive performance. However, RoBERTa has shown better results compared to other transformers based methods.

%%%%%%%%%%%    Table-3 %%%%%%%%%

\begin{table}[h]
\centering

\begin{tabular}{lcl}
\hline \textbf{Classifier} & \textbf{F1 score}  \\ 
\hline
BERT  & 0.88634  \\
ALBERT  & 0.87786 \\
DistilBERT & 0.88061  \\
RoBERTa & \textbf{0.88991} \\
\hline
\end{tabular}

\caption{\label{tab-3} F1 score of transformer-based approaches. }

\end{table}

% ////////////////////////////////////////////////////////////////////
\section{Conclusion}
\label{sec-7}

Classifying Twitter texts has been at the forefront of various NLP applications. Here, in this paper we have worked on one such task of classifying a tweet as informative or uninformative in the context of Covid-19. We have extensively compared the performance of various methods for this task. We applied conventional approaches and the latest state-of-the-art transformer-based methods. The results shows that the RoBERTa gives superior result on this task.

Since Twitter is primarily a microblogging media, short text classification using topic modeling \citep{zhang2013short, blei2009topic}, and topic-enhanced embedding-based approach. \citep{li2016tweetsift} can also be useful for tweet classification. In future work, we wish to apply topic modeling to produce enhanced word embeddings for this task. 

%We would also like to try out the use of metadata to investigative the role of biasness in the tweets for classification.

\bibliographystyle{acl_natbib}
\bibliography{anthology,emnlp2020}

\end{document}